\PassOptionsToPackage{unicode}{hyperref}
\PassOptionsToPackage{hyphens}{url}
\PassOptionsToPackage{dvipsnames,svgnames,x11names}{xcolor}
\documentclass[
runningheads]{llncs}
\usepackage{xcolor}
\usepackage{amsmath,amssymb}
\usepackage{iftex}
\usepackage{adjustbox}
\ifPDFTeX
\usepackage[T1]{fontenc}
\usepackage[utf8]{inputenc}
\usepackage{textcomp} 
\else 
\usepackage{unicode-math} 
\defaultfontfeatures{Scale=MatchLowercase}
\defaultfontfeatures[\rmfamily]{Ligatures=TeX,Scale=1}
\fi
\usepackage{lmodern}
\ifPDFTeX\else
\fi
\IfFileExists{upquote.sty}{\usepackage{upquote}}{}
\IfFileExists{microtype.sty}{
	\usepackage[]{microtype}
	\UseMicrotypeSet[protrusion]{basicmath} 
}{}
\makeatletter
\ifx\paragraph\undefined\else
\let\oldparagraph\paragraph
\renewcommand{\paragraph}{
	\@ifstar
	\xxxParagraphStar
	\xxxParagraphNoStar
}
\newcommand{\xxxParagraphStar}[1]{\oldparagraph*{#1}\mbox{}}
\newcommand{\xxxParagraphNoStar}[1]{\oldparagraph{#1}\mbox{}}
\fi
\ifx\subparagraph\undefined\else
\let\oldsubparagraph\subparagraph
\renewcommand{\subparagraph}{
	\@ifstar
	\xxxSubParagraphStar
	\xxxSubParagraphNoStar
}
\newcommand{\xxxSubParagraphStar}[1]{\oldsubparagraph*{#1}\mbox{}}
\newcommand{\xxxSubParagraphNoStar}[1]{\oldsubparagraph{#1}\mbox{}}
\fi
\makeatother

\usepackage{longtable,booktabs,array}
\usepackage{calc} 
\usepackage{etoolbox}
\makeatletter
\patchcmd\longtable{\par}{\if@noskipsec\mbox{}\fi\par}{}{}
\makeatother
\IfFileExists{footnotehyper.sty}{\usepackage{footnotehyper}}{\usepackage{footnote}}
\makesavenoteenv{longtable}
\usepackage{graphicx}
\makeatletter
\newsavebox\pandoc@box
\newcommand*\pandocbounded[1]{
	\sbox\pandoc@box{#1}%
	\Gscale@div\@tempa{\textheight}{\dimexpr\ht\pandoc@box+\dp\pandoc@box\relax}%
	\Gscale@div\@tempb{\linewidth}{\wd\pandoc@box}%
	\ifdim\@tempb\p@<\@tempa\p@\let\@tempa\@tempb\fi
	\ifdim\@tempa\p@<\p@\scalebox{\@tempa}{\usebox\pandoc@box}%
	\else\usebox{\pandoc@box}%
	\fi%
}
\def\fps@figure{htbp}
\makeatother

\usepackage[backend=biber, style=lncs]{biblatex}
\makeatletter
\@ifpackageloaded{caption}{}{\usepackage{caption}}
\AtBeginDocument{%
	\ifdefined\contentsname
	\renewcommand*\contentsname{Table of contents}
	\else
	\newcommand\contentsname{Table of contents}
	\fi
	\ifdefined\listfigurename
	\renewcommand*\listfigurename{List of Figures}
	\else
	\newcommand\listfigurename{List of Figures}
	\fi
	\ifdefined\listtablename
	\renewcommand*\listtablename{List of Tables}
	\else
	\newcommand\listtablename{List of Tables}
	\fi
	\ifdefined\figurename
	\renewcommand*\figurename{Figure}
	\else
	\newcommand\figurename{Figure}
	\fi
	\ifdefined\tablename
	\renewcommand*\tablename{Table}
	\else
	\newcommand\tablename{Table}
	\fi
}
\@ifpackageloaded{float}{}{\usepackage{float}}
\floatstyle{ruled}
\@ifundefined{c@chapter}{\newfloat{codelisting}{h}{lop}}{\newfloat{codelisting}{h}{lop}[chapter]}
\floatname{codelisting}{Listing}

\makeatother
\makeatletter
\@ifpackageloaded{caption}{}{\usepackage{caption}}
\@ifpackageloaded{subcaption}{}{\usepackage{subcaption}}
\makeatother
\makeatletter
\@ifpackageloaded{algorithm}{}{\usepackage{algorithm}}
\makeatother
\makeatletter
\@ifpackageloaded{algpseudocodex}{}{\usepackage[noEnd=false, indLines=false, italicComments=false, rightComments=false, commentColor=black, beginComment=//~]{algpseudocodex}}
\makeatother
\makeatletter
\@ifpackageloaded{caption}{}{\usepackage{caption}}
\makeatother
\makeatletter
\newcommand\fs@nocaption{
	\def\@fs@cfont{\bfseries}
	\let\@fs@capt\floatc@plain
	\def\@fs@pre{}%
	\def\@fs@post{\kern2pt\hrule}%
	\def\@fs@mid{\hrule\kern2pt}%
	\let\@fs@iftopcapt\iftrue}
\newcommand{\algpxEndIndentHelper}{\algpx@endIndent}
\newcommand{\algpxSetCommentColor}[1]{\renewcommand{\algpx@commentColor}{#1}}
\newcommand{\algpxSetBeginComment}[1]{\renewcommand{\algpx@beginComment}{#1}}
\newcommand{\algpxSetEndComment}[1]{\renewcommand{\algpx@endComment}{#1}}
\makeatother

\usepackage{bookmark}
\IfFileExists{xurl.sty}{\usepackage{xurl}}{} 
\makeatletter
\@ifundefined{xmpquote}{\newcommand{\xmpquote}[1]{#1}}{}
\makeatother
\hypersetup{
	pdftitle={AGrEe: a NeSy Automatic Gradient Engine},
	pdfauthor={Giacomo Bergami},
	pdfkeywords={\xmpquote{Tensor Algebra Optimization}, \xmpquote{Machine
			Learning}, \xmpquote{Einstein Sums}, \xmpquote{Tensor
			Kernels}, \xmpquote{SIMD}, \xmpquote{Data Parallelism}},
	colorlinks=true,
	linkcolor={blue},
	filecolor={Maroon},
	citecolor={Blue},
	urlcolor={blue},
	pdfcreator={LaTeX via pandoc}}

\title{AGrEe: a NeSy Automatic Gradient Engine}
\author{
	Giacomo Bergami\thanks{This publication is independent of the author's work at Brain Technologies srl. and does not represent the opinions of Brain Technologies srl }\inst{1}~\orcidID{0000-0002-1844-0851}}
\institute{
	Brain Technologies%
	, Turin%
	, Italy \email{bergamigiacomo@gmail.com}%
}
\begin{document}
	\maketitle
	
	\begin{abstract}
		The Automatic GRadient Engine (AGrEe) offers an adaptive pipeline
		seamlessly supporting both sparse and dense tensor inputs, while also
		providing tensors or loss functions as output layer nodes. By enhancing
		JIT-compile time and computation graph compression by reducing the
		computation node size as well as providing novel solutions to
		efficiently compute Einstein Sum queries, we offer a solution
		outperforming PyTorch even when JIT compiled over C++. We also offer
		ForESum, a learning-driven query plan algorithm using learning-based
		agents to estimate the best kernel composition to be picked to schedule.
	\end{abstract}
	
	\floatname{algorithm}{Algorithm}

	\section{Introduction}\label{introduction}
	
	Neuro-Symbolic Computing (NeSy) is an emergent framework for the
	artificial intelligence landscape reconciling the statistical robustness
	of connectionist architectures with the rigorous interpretability and
	compositionality of formal logic. While \emph{TensorLog}
	\cite{cohen2016tensorlog} translates Datalog programs into
	probabilistic factor graphs over numerical parameters, Differentiable
	Inductive Logic Programming (\(\partial\)ILP)
	\autocite{evans2018learning} relaxes discrete rule-selection mechanics
	into continuous optimisations over structural clauses. In real-world
	data scenarios, their associated tensor representations are extremely
	sparse and high dimensional. Current state-of-the-art AI computational
	frameworks are unequipped to support them efficiently. General-purpose
	deep learning libraries such as PyTorch \autocite{paszke2019pytorch}
	were primarily designed for the dense, uniform tensor operations typical
	of traditional deep learning, where sparse operations are mainly
	supported for low-rank tensors. Thus, dense Einstein Sums would require
	densifying the tensor and, by doing so, the operands might hardly fit on
	GPUs. Even specialised compilation frameworks designed specifically for
	sparse tensor algebra, such as the Tensor Algebra Compiler (TACO)
	\autocite{kjolstad2017taco}, are fundamentally restricted to static
	execution graphs and lack a native, robust automatic differentiation
	(autograd) engine.
	
	The database community has already adopted learned query optimisation by
	asymmetric \textbf{Teacher/Student architectures} by decoupling heavy,
	computationally intensive query plan exploration from the
	latency-critical online execution path , which can be further distilled
	into an ultra-low latency model for inferring at sub-two-microsecond
	regime. Still, state-of-the-art learned optimisers in the relational
	domain, such as Neo \autocite{marcus2019neo} and Balsa
	\autocite{marcus2021balsa}, operate under the assumption of stable
	relational schemas and static row dimensions. Conversely, tensorial
	workloads are heavily characterised by \emph{dynamic shapes} --- as
	caused by the incresed of the inferred facts and rules --- which
	fundamentally distort the geometric properties of the underlying
	computational graph at runtime. While relational cost models primarily
	optimise for logical metrics or disk input/output (I/O) boundaries,
	tensor compilation is intrinsically memory-bound, governed by the
	hardware-specific topologies of static random-access memory (SRAM),
	level-2 caches, and high-bandwidth memory (HBM). Despite these
	limitations, the AI ecosystem must critically assimilate foundational
	concepts from relational query optimisation to transcend the current
	limitations of greedy, heuristic-driven graph compilation.
	
	AGrEe solves some of these limitations through a unified framework
	providing full support for sparse data for specific NeSy frameworks
	requiring generalised einsum planning, end-to-end backpropagation,
	support of heterogeneous tensor formats (COOrdinate list format (COO),
	CSF, dense) as well as representing COO as streams to minimise
	intermediate tensor materialisation. The combined provision of better
	JIT graph compression and the presence of better Einstein Sum kernels,
	selectable through a learning-based algorithm (Forecasted Einstein
	Sums (ForESum)), enable our proposed framework to outperform PyTorch
	over the presented experiments.
	
	\section{Related Work}\label{related-work}
	
	TACO \autocite{kjolstad2017taco} natively unifies \texttt{einsum}
	planning and code generation through concrete index notation, merge
	lattices, and loop-nest scheduling. Unlike deep learning pipelines, TACO
	does not provide built-in reverse-mode backpropagation across arbitrary
	computational graphs, requiring external differentiable programming
	frameworks to synthesise sparse gradients. Its primary architectural
	strength lies in storage format generality, supporting arbitrary
	per-dimension combinations of dense, COO, and hierarchical CSF formats
	within single compound expressions. However, streaming execution is
	limited to single-kernel lexicographical co-iteration, necessitating
	specialised dataflow abstractions. TACO's cost model is restricted to
	local analytical loop-nest heuristics and schedules without a global,
	end-to-end graph cost objective. Given the several limitations and its
	inability to act as a fully-fledged automatic gradient engine, we remove
	this from our comparison as our system provides more capabilities than
	the ones provided by TACO. PyTorch and its TorchInductor compiler
	\autocite{ansel2024pytorch} delegate multi-operand \texttt{einsum} path
	selection to pre-lowering \texttt{opt\_einsum} heuristics
	\autocite{smith2018opt} before decomposing expressions into binary
	operations within a computation graph. Although PyTorch provides
	comprehensive dense reverse-mode differentiation through Autograd,
	sparse backpropagation is governed by constrained Vector-Jacobian
	Product (VJP) rules that require strict structural invariants such as
	coordinate coalescing. Across storage formats, the framework supports
	dense arrays alongside COO, Compressed Sparse Row (CSR), Compressed
	Sparse Column (CSC), and Block Sparse Row (BSR) layouts, yet lacks
	native support for arbitrary multi-dimensional CSF hierarchies.
	Streaming execution is likewise externalised to asynchronous host
	pipelines (\texttt{DataLoader}) rather than intra-graph dataflow
	scheduling, while its cost model relies on rule-based heuristic fusion
	paired with compile-time Triton kernel autotuning
	\autocite{ansel2024pytorch}. By comprising most of the desiderata, it is
	then benchmarked against AGrEe+ForESum.
	
	Notwithstanding the recent developments in AI, current query plan cost
	models fail to optimise over the aforementioned conditions due to three
	structural blind spots.
	
	\emph{First}, contraction path selection (einsum), reverse-mode AD graph
	generation, sparse layout assignment, and kernel fusion execute as
	isolated compiler passes. A contraction order chosen to minimise dense
	forward FLOPs frequently triggers catastrophic intermediate fill-in or
	forces expensive layout conversions during the backward pass. On the
	other hand, the proposed framework allows for supporting arbitrary data
	formats for both input and output, while also requiring specific data
	outputs for optimisation purposes, with the sole exception of the
	Einstein Sum optimised node ForESum, which is dictating the preferred
	output format depending on the task of interest.
	
	\emph{Second}, cost models rely on static tensor dimensions or known
	non-zero counts ((nnz)). Contracting two sparse CSF or COO tensors in an
	einsum---or computing Vector-Jacobian Products (VJPs) during
	backpropagation---produces intermediate tensors whose sparsity patterns
	depend on runtime coordinate intersections. Static cost models cannot
	accurately bound memory or compute costs without runtime cardinality
	estimation. ForESum attempts to solve this problem by exploiting
	Teaching/Student architectures with gated decisions, so as to ensure that
	the decision cost does not provide a significant computation overhead to
	the actual Einstein Sum execution.
	
	\emph{Last}, true pipeline streaming requires aligned lexicographical
	traversal across operators, while transitioning between unsorted COO,
	tree-compressed CSF, and dense vectors mid-pipeline requires blocking
	materialisation and radix-sorting passes. We attempt to solve the
	problem by first ensuring that our query plan can be represented as a
	DAG as well as considering pre-sorted COO, while providing efficient
	transformations between COO and CSF that might be implemented over the same
	vector allocated memory with minimal to zero reallocation costs.
	
	\section{AGrEe JIT compilation}\label{agree-jit-compilation}
	
	\emph{Despite our pipeline\footnote{\url{https://github.com/gyankos/AGREE}}
		supports nodes with backward passes, optimisation tapes and loss
		operators, these discussions are omitted due to space limitations as
		they highly overlap with current implementations. We limit our
		discussion of the JIT compiler algorithm over the pipeline computation
		graph.}
	
	At the time of writing, AGrEe JIT optimiser can be run only over
	acyclic graphs. \emph{First}, we annotate whether each node of the
	original computation graph can be retained as a sparse node: this might
	happen if either if the current node is an input tensor, or if this node
	provides a forward pass keeping zero cells at zero. This potentially
	generates a sparse operand mask \(M\), that is then provided to the
	ForESum node operator. \emph{Second}, we create affine operations
	clusters by merging together scalar algebra operations with an optional
	cell-wise tensor sum at the head for nodes left untouched; nodes are
	merged until the next node is no more a scalar algebra operation or the
	chain is broken by multiple outputs; this will lead to nodes having a
	forward pass of \(y = m\sum_ix_i  + c\), where slope \(m\) and bias
	\(c\) are derived from the accumulated operations. \emph{Third}, we
	merge the remaining nodes that can be refactored as \(y=x+\prod_i f_i\),
	thus considering binary CellWiseSums, where one of the operands is
	either a Hadamard Product, a product implementation of the and logic
	operator, thus including equivalent fuzzy logic operators. \emph{Last},
	over the remaining non-merged nodes, we scan the graph from the output
	nodes of a computation graph and find family chains to fuse: we fuse
	einsum nodes into one single node, while simplifying double logical
	negations into one single node. Furthermore, with respect to Einstein
	Sums, we early detect cases leading to trivial single-operand
	reductions, transposition, or diagonalisation, and rewrite those cases
	with the most appropriate unary operator. We extend our pre-processing
	to other families of computation, such as minimisation and maximisation
	operators. Clustered nodes are then merged according to a graph grouping
	operator \autocite{Junghanns}.
	
	Differently from other pipelines, a user might specify the desired
	\emph{delivery} format of the resulting tensors, used by the later
	gateway decision step. Such choice is then propagated backwards from the
	output nodes, where the propagation is blocked at input, which as at the
	discretion of the specific type of loaded tensors. Einstein Sum node
	operators forces an exception: as their query plan might override the
	choice of representation format, all the representation formats from
	their position onwards will be dictated by the query plan itself, while
	any further backpropagation of such suggestion is blocked.
	
	Sparse tensors, as the one stored in CSF \autocite{csf} or COO
	\autocite{coo} layouts, introduce non-uniform sparse data
	representations violating the locality principle and resorting to
	frequent random-access memory gathering \autocite{smith2015splatt}.
	While GPUs rely on strict memory contiguity and uniform execution paths,
	modern CPUs can parse complex metadata and perform bitwise operations
	without pipeline collapse \autocite{chambi2016roaring}, thus providing
	better adaptations to sparse tensor layouts. Furthermore, parallel
	accelerators possess overall significantly less physical memory than the
	host CPU, thus forcing bidirectional data migration across the interface
	bus when considering a huge volume of data. The latency of this
	host-to-device communication routinely dwarfs the actual computational
	execution time of the kernel
	\autocite{intel2024optimization,kjolstad2017taco}. By shifting the
	design focus toward optimising sequential host-side operations, we first
	ensure that when data must reside in or transit through system memory,
	the core utilises the maximum physical bandwidth of the single-core
	memory channel (e.g., via non-temporal or sequential write-once paths),
	bypassing the cross-device transport tax entirely. Given this, the
	current pipeline setup prefers sequential kernels on CPU and currently
	implements no GPU tensor representation. Notwithstanding the former, the
	implementative separation between backend representation, computation
	graph nodes, and associated kernels allows our architecture to be
	seamlessly extended in the near future to also support other execution
	models and tensor representations. CPU-bounded experiments in
	Section~\ref{sec-expfirst} suggest that such optimisation improves over
	PyTorch runtime by providing a more compact computation graph.
	
	\section{ForESum
		Algorithm}\label{b58fc729-690b-4000-b19f-365a4093b2ff7b7b3c2061637220466f724553756d203e7d7d-algorithm}
		
			\floatstyle{ruled}
		\restylefloat{algorithm}
		\begin{algorithm}[htb!]
			\caption{\label{alg-foresum}Forecasted Einstein Sums algorithm (ForESum)}
			\begin{algorithmic}[1]
				\setbool{algpx@italicComments}{true}%
				\setbool{algpx@rightComments}{true}%
				\algpxSetCommentColor{blue}%
				\algpxSetBeginComment{$\rhd$~}%
				\Procedure{ForESum}{$op_1,\dots,op_n;\; $EinsumStr, $M$, Engine=auto} \Comment{$n\geq 2$}
				\State $\vec{x}\gets $\Call{Features}{$op_1,\dots,op_n;\; $EinsumStr}; $E\gets $EinsumStr
				\State $C\gets \{1\leq k\leq n|(\text{sparse}(op_k) \vee (k\in M \wedge \text{numel}_k>0))\wedge (\text{nnz}_k/_0\text{numel}_k\leq 0.5) \}$
				\If{\Call{Gate}{$\vec{x}$}$=$Routes} \Comment{Exp., Sec. \ref{sec-gateffect}}\label{line-gate}
				\State $f\gets$\Call{Class}{E}
				\If{$x[\text{dflop}]< \theta_f$} \label{threshold-gate}
				\If{$f\neq$Other \textbf{and} \textit{delivery}=dense}
				\State $\hat r=\arg\min_{r\in\mathcal{R}_f\cap\mathcal{A}(E,\vec{x})}G_{fr}(\vec u)$ \label{gatewayact} \Comment{Gateway decision}
				\EndIf
				\Else
				\State $R\gets\{r_i\in S(\vec{x})|i\leq 3\wedge\mathbb{P}(r_i\textrm{ is }\texttt{ok}|\vec{x})\geq 0.5\}$ \Comment{Student top-3 filtering}\label{student3}
				\State $\hat r=\mathrm{argmin}_{r\in R,\mathbb{P}(r\textrm{\,is\,}\mathtt{ok}|\vec{x})\geq 0.5}=\hat{t}_{r;1.}(\vec{x})$ \Comment{$\leq$3-Expert decision} \label{forexpert}
				\EndIf
				\State \Return \textbf{if} $\hat r\neq$\texttt{null} \textbf{then}  $\hat r(op_1,\dots,op_n)$ \textbf{else} \Call{CmpFS}{$op_1,\dots,op_n$; E} \label{fallooff}
				\Else \,\textbf{if} \Call{Gate}{$\vec{x}$}$=$Dense \textbf{then} $C\gets\emptyset$
				\EndIf
				\State $dop_k\gets $\Call{densify}{$op_k$} \textbf{for each} $k$ \textbf{s.t.} sparse($op_k$)
				\If{EinsumStr is star \textbf{and} $C=\emptyset$} \Return \Call{FusedSt}{$op_1,\dots,op_n;\; $EinsumStr}
				\Else\, \Return \Call{FallBackEinsum}{$op_1,\dots,op_n;\; $EinsumStr, $C$}
				\EndIf
				\EndProcedure
			\end{algorithmic}
		\end{algorithm}
		\floatstyle{plain}
	
	The ForESum Algorithm (Algorithm~\ref{alg-foresum}) relies on query plan
	supporting novel kernels outperforming current handcrafted solutions
	implementable over AI pipeline libraries. By shifting the design focus
	toward optimising sequential host-side operations, we opt for data-level
	parallelisation through portable SIMD abstractions such as the one
	provided by Google Highway \autocite{GoogleHighway}, with the added
	benefit of achieving execution times bounded in the micro-second regime
	for sparse and smaller tensors. Similar solutions can also be adopted
	when single operators can be fully stored within a single page in
	memory, thus minimising the overall number of page faults. Due to space
	limitations, we omit discussing the implementation of specific Einstein
	Sum kernels leading to overall scalability improvements. Still,
	Section~\ref{sec-expkernel} provides experimental evidence that the
	provided routes outperform state-of-the-art solutions available in
	current AI pipeline libraries. We limit our discussion to the training
	of the cost model (Section~\ref{sec-traincost}) and to showing how such
	pipeline is used to run Einstein Sums (Section~\ref{sec-actforesum})

	\subsection{Training the Cost Selector Stack}\label{sec-traincost}
	
	The trained \textbf{gate} (L. \ref{line-gate}) decides whether to attempt scheduling
	one of the learned routes (Routes) or deciding to fall back to either
	dense or legacy plans by estimating their running times with basic
	linear models, \(\hat{t}_R\), \(\hat{t}_D\), and \(\hat{t}_L\)
	respectively. Its training samples \(i\) are einsum queries generated
	from a grid of workload families (SpMM, chain, star, batched), sizes,
	densities, seeds and directions. As part of the \(\vec{x}\) features,
	such gate notably considers whether the sparse operands are already
	bound as sparse, and the cost estimates given the operands' sizes. Each
	alternative gets a time prediction that is linear in non-negative
	coefficients. The coefficients of each alternative are fitted by
	non-negative weighted least squares on relative error:
	\[\min_{\theta_c\ge 0}\ \sum_i w_i\,\frac{(\hat t_{i,c}-t_{i,c})^2}{t_{i,c}^2},\qquad w_i=1+4\exp\!\big(-2\,|\ln (t_{i,(2)}/t_{i,(1)})|\big),\]
	where \(t_{i,(1)}\le t_{i,(2)}\) are the two smallest measured times of
	sample \(i\). The weight \(w_i\) emphasises samples whose two best
	alternatives are close, since those decide the regret. The model's
	argmin is corrected by a decision tree used for the final decision of
	maximum depth of 5 on the augmented vector
	\(\vec z_i=(\vec x_i,\hat t_{iL},\hat t_{iD},\hat t_{iR})\). A leaf
	\(\ell\) is labelled with the alternative of shortest total measured
	time over its samples and splits are chosen to minimise the regret with
	respect to the per-sample oracle.
	
	A subsequent \textbf{threshold gate} (L. \ref{threshold-gate}) lets the
	default planner run when the expert/student decision time \(\omega_f\)
	might dominate the actual kernel computation for a set of operands \(c\)
	considering a specific query-kernel type \(f\) where \(a_c\) and \(b_c\)
	respectively provide the default plan and best-case running time
	scenarios, for which we derive a gain \(g_c=a_c-b_c-\omega_f\). Given
	\(\text{dflop}_c\) the full iteration space of the native loop nest
	providing an upper bound for the maximum number of expected floating
	point arithmetic operations, the desired threshold value for the
	specific family \(f\) is the candidate \(\theta_f\) minimising
	\(\sum_{c,\text{dflop}_c<\theta_f}g_c\).
	
	The \textbf{Expert Model} (L. \ref{forexpert}) is trained to both estimate the expected route
	running times as well as the ok/failures while executing the routes. For
	doing this, we collect data features describing both the system where it is
	running on, input tensor information, and expected output tensor, as
	well as features describing the algorithm itself. We denote these as
	\(\vec{x}\). For every expert, meaning one (algorithm, forward or
	backward, \emph{delivery}) triple as a route \(r\), there are two
	forests of 30 CART trees (sklearn random forests, leaf size 3, at most
	64 leaf nodes): the regression prediction\footnote{\(\operatorname{UCB}_{\mathcal{F}}(x; \lambda) = \mu_{\mathcal{F}}(x) + \lambda \sigma_{\mathcal{F}}(x)\)}
	in milliseconds
	\(\hat{t}_{r;\lambda}(\vec{x})=(\mathrm{exp}(UCB_{\{ET_{ir}\}_{i=1}^{30}}(\vec{x},\lambda))-1)\cdot 10^{-3}\)
	and a binary \texttt{ok} classifier
	\(\mathbb{P}(r\textrm{\,is\,}\texttt{ok}|\vec{x})=\mathbb{E}_{1\leq i\leq 30}[{E\mathtt{ok}}_{ir}(\vec{x})]\).
	In the former, the trees are fitted on \(\ln(1+1000t_{ms})\) over the
	\texttt{ok} routes; in the latter, \texttt{ok} is set to false when the
	probe refuses to run the query over the kernel, if the tests lead to
	out-of-memory exceptions, or whether this case is intercepted by the
	memory gate refusing the case.
	
	The ranking-based \textbf{Student Model} (L. \ref{student3}) is a 3-layer sequential
	Multi-Layer Perceptron (MLP) trained via a ListNet listwise loss. It
	uses a subset of rank \(D\) features as the expert model. First, for
	each sample \(i\in S\) and route \(r\), Ground-truth classification
	target times \(T_{ir}\) are constructed from measured execution times
	\(t_{ir}\) in milliseconds as \(\ln (\max (t_{ir},10^{-6}))\); we also
	map the \texttt{ok} scores into a binary matrix \(O_{ir}\), and restrict
	training over instances with \(O_{ir}=1\). Second, the input is
	normalised: given \(\vec{\mu}\) and \(\vec{\sigma}\) the feature-wise
	mean and standard deviation computed over the training set, the student
	model is trained over a dataset transformed with signed-log for
	compressing heavy-tailed distributions and extreme magnitudes within the
	input data while preserving the sign of the features:
	\[\overline{\vec{x}_{i}}=\frac{(\mathrm{sgn}(\vec{x}_{i})\odot\ln(1+|\mathbf{x}_{i}|))-\vec{\mu}}{\vec{\sigma}+\epsilon}\]
	
	Third, the normalised features are fed to the aforementioned MLP model
	\((d_0, d_1, d_2, d_3) = (D, 32, 32, |R|)\) with \(D\) denoting the
	expert model's features:
	
	\[f_{\theta }(\vec{x}_i)=\mathbf{W}_{3}\,\mathrm{SiLU}\!\left(\mathbf{W}_{2}\,\mathrm{SiLU}(\mathbf{W}_{1}\vec{x}_i+\mathbf{b}_{1})+\mathbf{b}_{2}\right)+\mathbf{b}_{3},\quad \theta =\left\{(\mathbf{W}_{\ell },\mathbf{b}_{\ell })\in \mathbb{R}^{d_{\ell }\times d_{\ell -1}}\times \mathbb{R}^{d_{\ell }}\right\}_{\ell =1}^{3}\]
	The resulting predicted utility vector is then transformed into a
	predicted probability distribution \(Q_{\theta,i}\) over the same valid
	subset of routes using a masked softmax:
	\[Q_\theta(r|\vec{x}_i)=\frac{O_{ir}\exp (f_{\theta }(\vec{x}_{i})_r )}{\sum _{k=1}^{|R|}O_{ik}\exp (f_{\theta }(\vec{x}_{i})_r)}\]
	
	We convert execution times into a target probability distribution
	\(P_i \in \Delta^{|R|-1}\) over valid routes by negating the log-times
	(so shorter times yield higher scores), scales them by a temperature
	hyperparameter \(\tau=0.3\), and applies a masked softmax, thus obtaining
	the following ground truth distribution:
	\[P_{ir}=\frac{O_{ir}\exp (-T_{ir}/\tau )}{\sum _{k=1}^{|R|}O_{ik}\exp (-T_{ik}/\tau )}\]
	
	The student model is then trained by minimising the ListNet
	cross-entropy loss \autocite{10.1145/1273496.1273513} between the ground
	truth distribution \(P_{i,r}\) and the predicted distribution
	\(Q_\theta(r|\vec{x}_i)\) over mini-batches \(\mathcal{B}\) which, in
	our experiments, have cardinality of 128. \(Q\) is not needed for
	inference, as we top-\(k\) routes ranking could be operated over the
	output logits as \(S(\vec{x})=\mathrm{argsort}_k(f_\theta(\vec{x}))\).
	
	Finally, the \textbf{gateway} (L. \ref{gatewayact}) replaces the full route decision \(r\) for
	\emph{small} operands \(c\), where evaluating the student and the
	experts would cost more than running the kernel. A set of operands
	belong to a small tier when its \(\text{dflop}_c\) is below a
	class-specific threshold \(\tau_f\), fitted on the training split so
	that the full decision is paid only where it pays off. The features in
	\(\vec{x}\) being considered at this stage are restricted to dflop,
	density, output cells upper bounds, and number of dense and sparse
	operands. The target is the log of the measured delivered time
	\(\ln t_{cr}\) of each successful route or kernel \(r\), which is then
	used to fit a simple linear model \(G_{fr}(\vec{x})\) which parameters
	are obtained by ridge regression on the small cases of the training
	split only with the intercept unpenalised, and only for routes with at
	least six training points:
	\[\min_\theta\ \sum_i\big(\ln t_{fr}-G_{fr}(\vec x_i)\big)^2+\lambda\|\theta_{-0}\|^2,\qquad \lambda=10^{-2},\]
	
	\subsection{ForESum Algorithm}\label{sec-actforesum}
	
	Algorithm~\ref{alg-foresum} orchestrates the aforementioned pipeline
	into a cohesive framework, where cases \(n\leq 1\) are handled at JIT
	compilation time as per the previous section. We denote \(/_0\) as the safe
	division, nnz as the number of non-zero elements in a tensor, numel the
	product of all the dimensions, \(\mathcal{R}_f\) as the set of all
	possible kernels applicable to a specific class of Einstein Sum queries
	\(f\), and \(\mathcal{A}\) as the set of applicable routes while also
	considering the operands and their representation. For small operands
	for which the dense delivery does not impact the overall runtime,
	we exploit the kernel route suggested by the gateway linear model (L.
	\ref{gatewayact}); otherwise, we let the student pre-filter the three
	possible experts (L. \ref{student3}) delivering the final route decision
	(L. \ref{forexpert}). If this final decision is not inconclusive, then
	we run the selected kernel-route and, otherwise, we let a Selinger-style
	optimiser combine some of the available routes using the trained cost
	models for the heuristic query plan decision (L. \ref{fallooff}). We
	consider the following fallback measures as a last resort:
	\textsc{FallBackEinsum}, using the Selinger-style optimizer
	\autocite{Selinger} with opt-einsum optimisations
	\autocite{smith2018opt} and potentially falling back to sparse path
	joins for the tensors indicated in \(C\) \autocite{assembleAndCompute},
	and \textsc{FusedSt}, using a similar approach to relational databases
	for aggregate push-down and factorised evaluation
	\autocite{10.1145/3003665.3003667}.

	\section{Experimental Results}\label{experimental-results}
	
	In AGrEe, we consider both our re-designed CPU tensor representation as
	well as a native C++ implementation of Python's numpy library (NumCpp)
	as two distinct backends. Optimal Python competitors vary depending on
	the experiment and are as indicated. Due to the lack of space, we refer
	the reader to our aforementioned GitHub repository and
	dataset\footnote{\url{https://osf.io/w5sh8/overview}} for further
	details on the different pipeline set-ups of our experiments, alongside
	with a more detailed analysis of our results. Except for the first
	experiment, we consider different classes of queries: \texttt{outer2}
	(\emph{ab,cd-\textgreater abcd}), \texttt{batched}
	(\emph{bij,bjk-\textgreater bik}), \texttt{star}
	(\emph{sa,sb,sc-\textgreater s}), \texttt{chain3}
	(\emph{bi,ia,ac-\textgreater bc}), and \texttt{triangle}
	(\emph{ij,jk,ji-\textgreater{}}) for the precise purpose of triggering
	different set of routes: spmm-like, dense-factor chain, multiway
	intersection, cyclic joins, and fall-back plans. Experiments were run on
	a Lenovo ThinkPad P16 Intel® Core™ Ultra 9 185H × 22 with 64GB of RAM.
	
	\subsection{Non-Einstein Sum Pipeline with Dense
		operands}\label{sec-expfirst}
	
\begin{figure}
		
		\centering{
			
			\includegraphics[width=1\linewidth,height=\textheight,keepaspectratio]{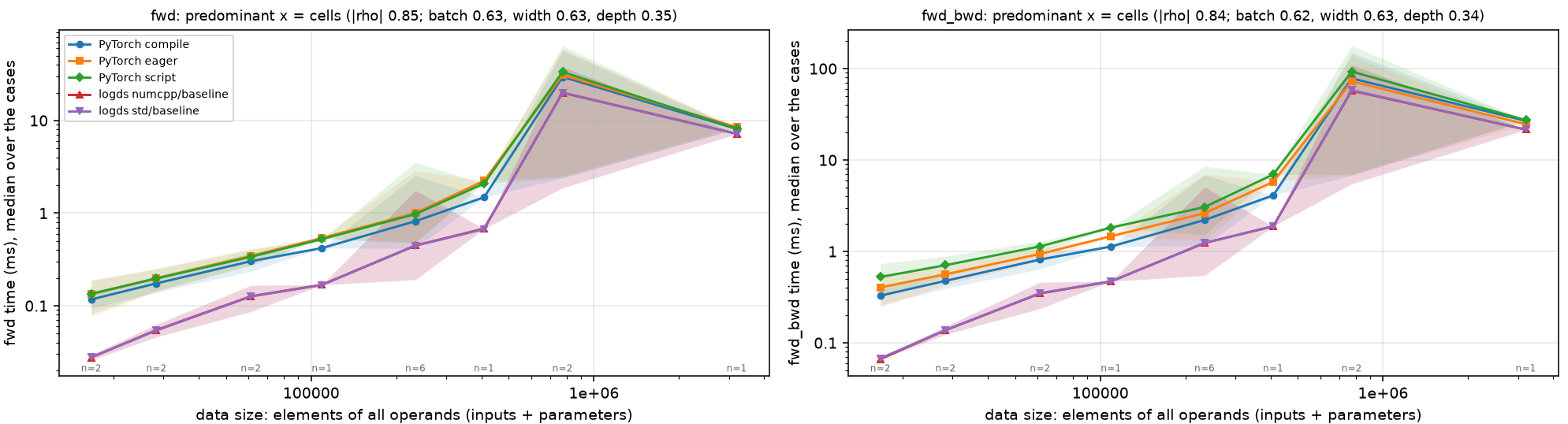}
			
		}
		
		\caption{\label{fig-elephant}Scalability plot for the first experiment
			with Forward and Backwards passes and basic GEMM operations.}
		
	\end{figure}

	To remark the versatility of the pipeline in terms of both considering
	dense and sparse operands, in opposition to PyTorch's mainly dealing
	with dense ones, we compare our pipeline as well as PyTorch by
	considering tensors being given as an input as fully dense operands. We
	consider the following PyTorch baselines: \emph{eager} is the baseline
	without JIT compilation, \emph{script} compiles the JIT graph without
	any node fusion optimisations while interpreting the graph in Python,
	while \emph{compile} compiles the JIT graph in C++ and runs the pipeline
	over fused nodes. We consider no computation of a loss function, but
	still consider the retropropagation cost by considering a fixed loss
	value that is then back-propagated. Figure~\ref{fig-elephant} summarises
	the findings in scalability results, showing that our solution constantly
	outperforms PyTorch solutions discussed above.
	
	PyTorch has two inefficiencies. First, there is a fixed cost per
	operation (dispatch, temporaries, autograd bookkeeping) dominating at
	small sizes while being averaged out at large ones. Second, each
	element-wise operation is a separate pass over the activations is the
	51\% of PyTorch's time outside GEMM at batch 8192. So at large sizes the
	remaining difference is data-dependent. Unlike PyTorch, our
	solution provides better compression on the JIT graph by fusing more
	operations, so as to potentially reduce numerical errors as well
	as computational overhead due to the creation of intermediate tensors.
	While PyTorch solution works on 11 nodes, our compaction reaches 6. From
	the depth sweep, a layer costs about 69 µs in PyTorch eager and about 21
	µs in logds, while the GEMMs in it are under 10 µs at the rate measured
	in section 2. Most of PyTorch's time at these sizes is dispatch,
	temporaries and autograd bookkeeping, not arithmetic. As PyTorch spends
	32.2 ms (51\%) outside GEMM and AGrEe 4.1 ms (11\%), that difference is
	the fusion saving.
	
	For larger datasets, the saving grows with size, and the ratio shrinks
	because the shared term grows faster. For a batch dimension of 8192, the
	total running time is 36.1 ms for the best AGrEe backend vs.~63.8 ms for
	the best PyTorch. In this, we remove 28 of PyTorch's 32 ms outside GEMM,
	which is our largest absolute saving across the board. As both PyTorch
	and AGrEe take around 32.0 ms for GEMM-related computations. In fact,
	for batch dimension of 8, the GEMM is negligible, so the ratio is almost
	purely ``other against other'' (197 µs against 46 µs, or 4.3×), even
	though the absolute saving is only 0.15 ms.
	
	On dense models the advantage is bounded by the shared GEMM, and the
	route selection of ForESum is what lifts that bound when sparsity is
	present. The following experiments further investigate this preliminary
	research hypothesis.
	
	\subsection{ForESum routes vs SciPy and PyTorch}\label{sec-expkernel}

	\begin{longtable}[]{@{}
			>{\raggedright\arraybackslash}p{(\linewidth - 12\tabcolsep) * \real{0.0962}}
			>{\raggedright\arraybackslash}p{(\linewidth - 12\tabcolsep) * \real{0.3}}
			>{\raggedright\arraybackslash}p{(\linewidth - 12\tabcolsep) * \real{0.1}}
			>{\raggedright\arraybackslash}p{(\linewidth - 12\tabcolsep) * \real{0.0962}}
			>{\raggedright\arraybackslash}p{(\linewidth - 12\tabcolsep) * \real{0.1058}}
			>{\raggedright\arraybackslash}p{(\linewidth - 12\tabcolsep) * \real{0.0865}}
			>{\raggedright\arraybackslash}p{(\linewidth - 12\tabcolsep) * \real{0.2115}}@{}}
		\caption{Comparing best ForESum routes with PyTorch and SciPy solutions
			for Einstein Sum kernels}\label{tbl-comparison}\tabularnewline
		\toprule\noalign{}
		\begin{minipage}[b]{\linewidth}\raggedright
			query
		\end{minipage} & \begin{minipage}[b]{\linewidth}\raggedright
			winning route (wr)
		\end{minipage} & \begin{minipage}[b]{\linewidth}\raggedright
			cases (share)
		\end{minipage} & \begin{minipage}[b]{\linewidth}\raggedright
			wr time
		\end{minipage} & \begin{minipage}[b]{\linewidth}\raggedright
			best Python (time)
		\end{minipage} & \begin{minipage}[b]{\linewidth}\raggedright
			speed-up
		\end{minipage} & \begin{minipage}[b]{\linewidth}\raggedright
			best Python
		\end{minipage} \\
		\midrule\noalign{}
		\endfirsthead
		\toprule\noalign{}
		\begin{minipage}[b]{\linewidth}\raggedright
			query
		\end{minipage} & \begin{minipage}[b]{\linewidth}\raggedright
			winning route (wr)
		\end{minipage} & \begin{minipage}[b]{\linewidth}\raggedright
			cases (share)
		\end{minipage} & \begin{minipage}[b]{\linewidth}\raggedright
			wr time
		\end{minipage} & \begin{minipage}[b]{\linewidth}\raggedright
			best Python (time)
		\end{minipage} & \begin{minipage}[b]{\linewidth}\raggedright
			speed-up
		\end{minipage} & \begin{minipage}[b]{\linewidth}\raggedright
			best Python
		\end{minipage} \\
		\midrule\noalign{}
		\endhead
		\bottomrule\noalign{}
		\endlastfoot
		triangle & \texttt{plan.join.dense} & 26\% & 18 µs & 302 us & 21.3x &
		\texttt{scipy\_csr} \\
		& \texttt{ndvec.einsum} & 25\% & 8.4 µs & 53 µs & 7.03x &
		\texttt{torch.einsum} \\
		& \texttt{legacy.dense} & 19\% & 4.1 µs & 33 µs & 8.57x &
		\texttt{torch.einsum} \\
		chain3 & \texttt{rows.chain} & 73\% & 0.5 µs & 8.3 µs & 24.9x &
		\texttt{torch\_dense} \\
		& \texttt{plan.mkl\_spmm.dense} & 7\% & 3.5 µs & 8.0 µs & 2.22x &
		\texttt{torch\_dense} \\
		& \texttt{plan.pairwise.dense} & 6\% & 3.0 µs & 8.8 µs & 2.76x &
		\texttt{torch\_dense} \\
		batched & \texttt{.own.rowsparse} & 27\% & 0.9 µs & 19 µs & 28.1x &
		\texttt{torch\_dense} \\
		& \texttt{.mkl.rowsparse} & 23\% & 3.6 µs & 165 µs & 67.4x &
		\texttt{scipy\_csr} \\
		& \texttt{.own.dense} & 22\% & 0.8 µs & 6.1 µs & 8.15x &
		\texttt{torch\_dense} \\
		star & \texttt{.keys} & 76\% & 1.1 µs & 27 µs & 37.5x &
		\texttt{torch\_dense} \\
		& \texttt{dfused..dense} & 17\% & 0.7 µs & 18 µs & 26.1x &
		\texttt{torch\_dense} \\
		& \texttt{.bitmap} & 4\% & 1.4 µs & 20 µs & 15.3x &
		\texttt{torch\_dense} \\
		outer2 & \texttt{.emit} & 75\% & 0.5 µs & 126 µs & 337x &
		\texttt{torch\_dense} \\
		& \texttt{.dense} & 23\% & 0.3 us & 9.6 µs & 13.1x &
		\texttt{torch\_dense} \\
		& \texttt{legacy.ndvec.dense} & 2\% & 647 µs & 1.2 ms & 1.84x &
		\texttt{torch.einsum} \\
	\end{longtable}
	
	This set of experiments focuses on the efficiency of the specific routes
	implemented in ForESum against ad hoc Python implementations used by
	common AI libraries. Table~\ref{tbl-comparison} provides a fair
	comparison between every single route from ForESum with PyTorch and
	re-implementation of sparse tensor operations over CSR representations
	over SciPy, which are mainly outperforming PyTorch solution for large
	sparse matrices. In particular, \texttt{torch\_dense} refers to
	re-implementations of the corresponding Einstein Sum query using dense
	PyTorch tensors, \texttt{scipy\_csr} refers to using SciPy sparse
	tensors operations over CSR tensor layouts, thus including explicit
	Kroneker Products, while \texttt{torch\_einsum} refers to exploiting
	explicit PyTorch matrix multiplication operations supporting sparse
	tensors, thus avoiding unfair comparison with the actual \texttt{einsum}
	operator strictly requiring that all operands must be dense. Please
	consider that all these Python libraries have C/C++ implementations for
	such CPU-based operations for efficiency purposes, thus providing
	overall fair comparisons with our pipeline.
	
	For \texttt{star} queries, our kernels avoid the creation of
	intermediate tensors by firstly summing up over the internal indices for
	then computing index-level intersection for matching indices within the
	Einstein Sum before emitting the actual result. On the other hand, the
	best solution over dense path scatters the COO operands into dense
	tensors and does the full multiply-accumulate. Similar considerations
	can be also drawn for \texttt{chain} queries, where competing solution
	almost fall back into dense tensor representations while our kernels
	scan row indices and avoid emitting empty rows. Similar considerations
	can be provided for \texttt{outer} queries, where the Kronecker product
	\autocite{DBLP:conf/edbt/BergamiMM17} directly yields the final result
	through sorted index scanning for immediately accessing the cells to be
	summed togehter using leap scans \autocite{info14030173}, while python
	solutions must materialise the dense \(n^r\) outer product, thus
	justifying the reported results.
	
	ForESum solutions for \texttt{batched} queries compact live rows into
	one CSR block using SIMD data processing and using caching to avoid
	costly re-allocations. On the other hand, dense binary matrix
	multiplications process every row, thus including empty ones, which
	explains our 28--67× figures. ForESum achieves the smallest gain where
	all solutions fall back to dense BLAS/GEMM-type kernel within
	\texttt{chain3}, \texttt{batched}, and \texttt{triangle} query plans.
	When the best solution is still not a native one enabling sparse
	tensors, the additional costs are roughly 8--10 µs for
	\texttt{torch.einsum} having additional densification costs.
	
	\subsection{ForESum Ablation
		Study}\label{b58fc729-690b-4000-b19f-365a4093b2ff7b7b3c2061637220466f724553756d203e7d7d-ablation-study}
	
	We now conduct an ablation study over the ForESum operands, where we
	compute the ratio between the overall model running time -- including
	the AI \emph{decision} for the route to be scheduled, computing the
	\emph{kernel}, and returning the final tensor over a given desired
	format (\emph{delivery}) -- and the best overall running time.
	
	We consider the following modes: the \textbf{Baseline} forces the execution of the \textsc{CmpFS} fallback by forcing all the $\theta_f$ score values to be too small to decide and forcing the class to \textsc{Other}. As a cost function for deciding over the kernels/routes to be picked, we consider simplistic time-complexity driven cost functions from the literature. \textbf{Trivial Cost} refers to
	having one single symbolic regression model for predicting the runtime
	given the local machine (e.g., last level cache size) as well as tensor
	(e.g., number of overall non-zero, operand dimension size) and trained
	through least squares on relative error thus replacing the student/expert model with a more simplistic variant.  The basic
	\textbf{ForESum-expert} considers the aforementioned expert CART models to
	pick the best route, while \textbf{ForESum-student} model only considers
	the student model as a valid top-1 prediction. With
	\textbf{ForESum+student+ok}, the model falls back to the expert decision
	when the picked route makes the expert model classify it as not
	\texttt{ok}. \textbf{+Gateway} indicates the activation piecewise-linear
	gateway overriding the full model decision over smaller datasets.
	
	\begin{figure}
		
		\centering{
			
			\includegraphics[width=1\linewidth,height=\textheight,keepaspectratio]{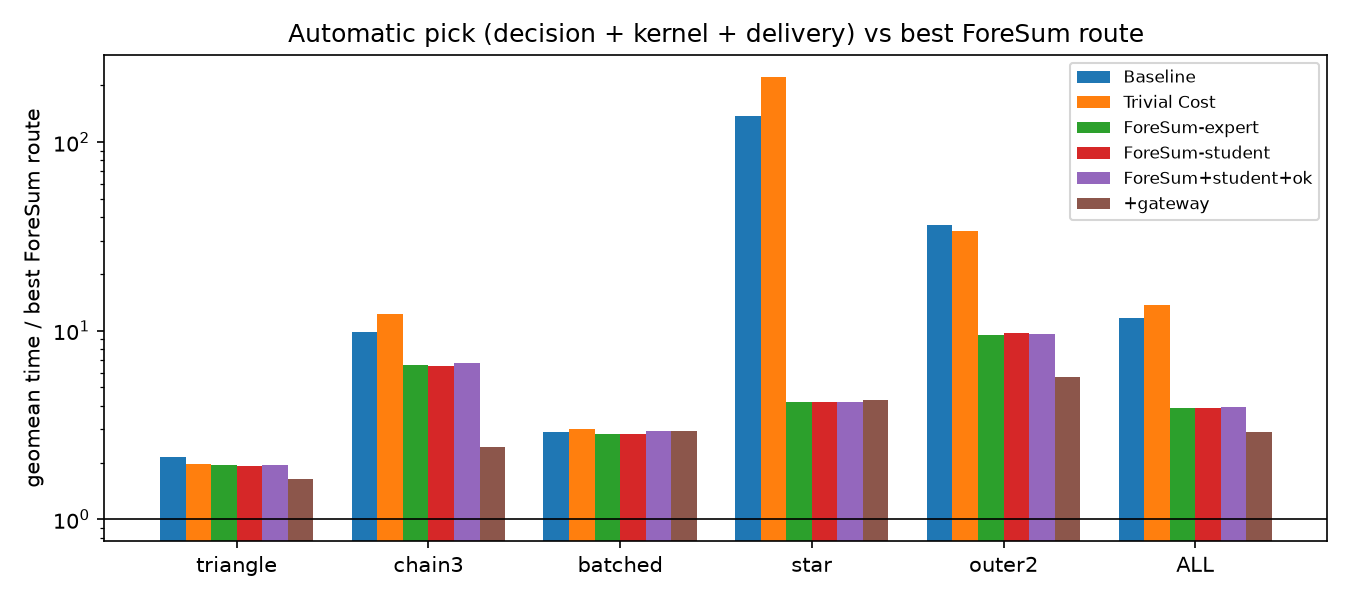}
			
		}
		
		\caption{\label{fig-ablation}Ablation study for the ForESum
			learning-based cost model.}
		
	\end{figure}%
	
	Figure~\ref{fig-ablation} shows that having a trivial linear regression
	model leads to mis-picks if compared to the proposed ForESum
	architecture, which on average leads to better route decisions leading
	to time minimisation. The ablation study also remarks that the, for the
	problem of interest, the student model behaves similarly to the
	full-expert model, thus providing optimal compression for the decision
	process as it avoids costly testing through tree forests. We also remark
	that the adoption of the gateway enables contained running time
	improvements when triggered. Overall, this study remarks the
	effectiveness of the proposed solution if compared to trivial solutions.
	The last set of experiments will remark the effect of the pre-gate on
	the overall time decision process while considering our solution as the
	aforementioned \textbf{+Gateway} variant.
	
	\subsection{Forward and Forward+Backward pipeline passes}\label{sec-gateffect}
	
	\begin{figure}
		
		\begin{minipage}[t]{\linewidth}
			
			\centering{
				
				\pandocbounded{\includegraphics[keepaspectratio]{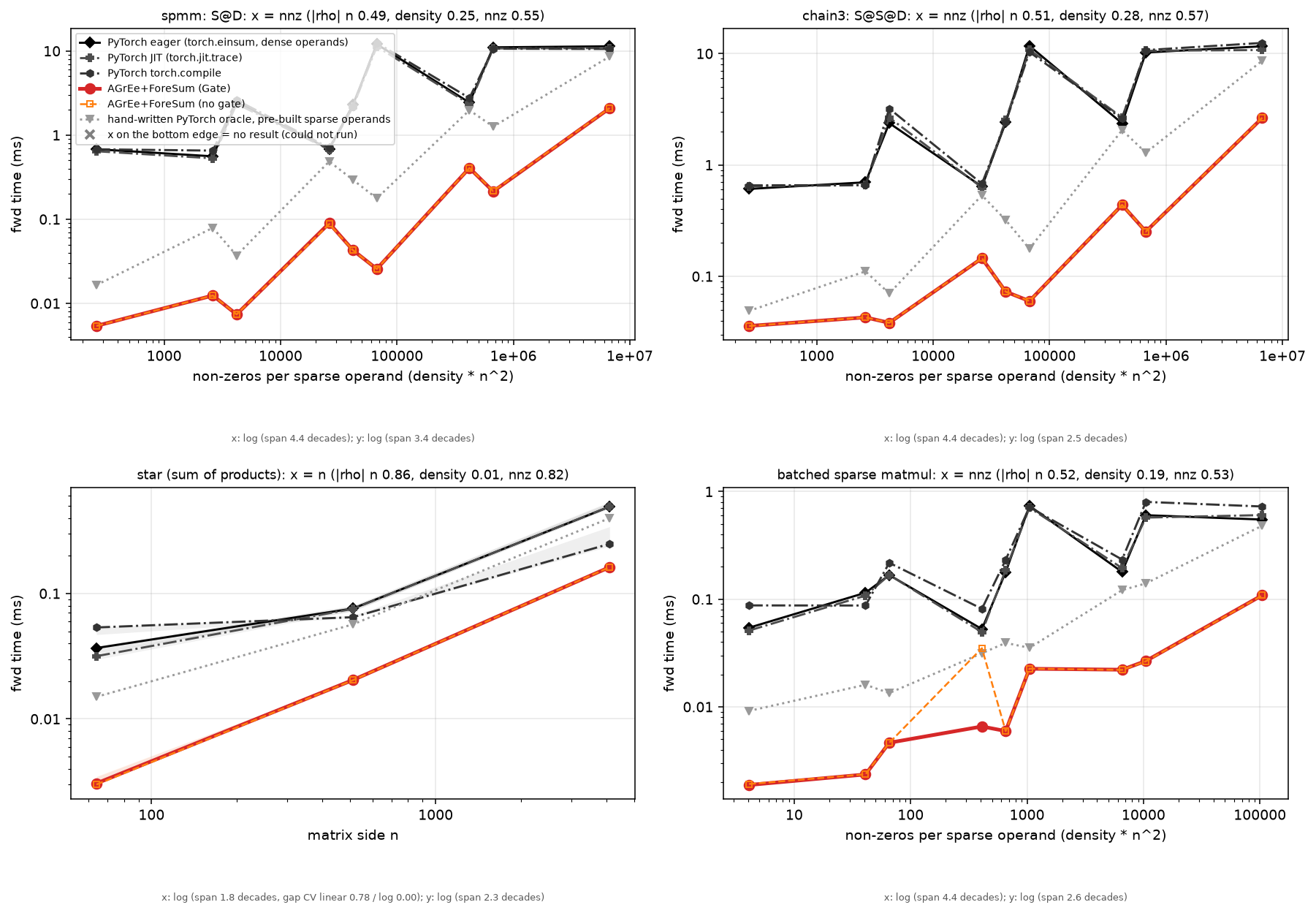}}
				
			}
			
			\subcaption{\label{fig-prima}Scalability results for just the forward
				step.}
			
		\end{minipage}%
		\newline
		\begin{minipage}[t]{\linewidth}
			
			\centering{
				
				\pandocbounded{\includegraphics[keepaspectratio]{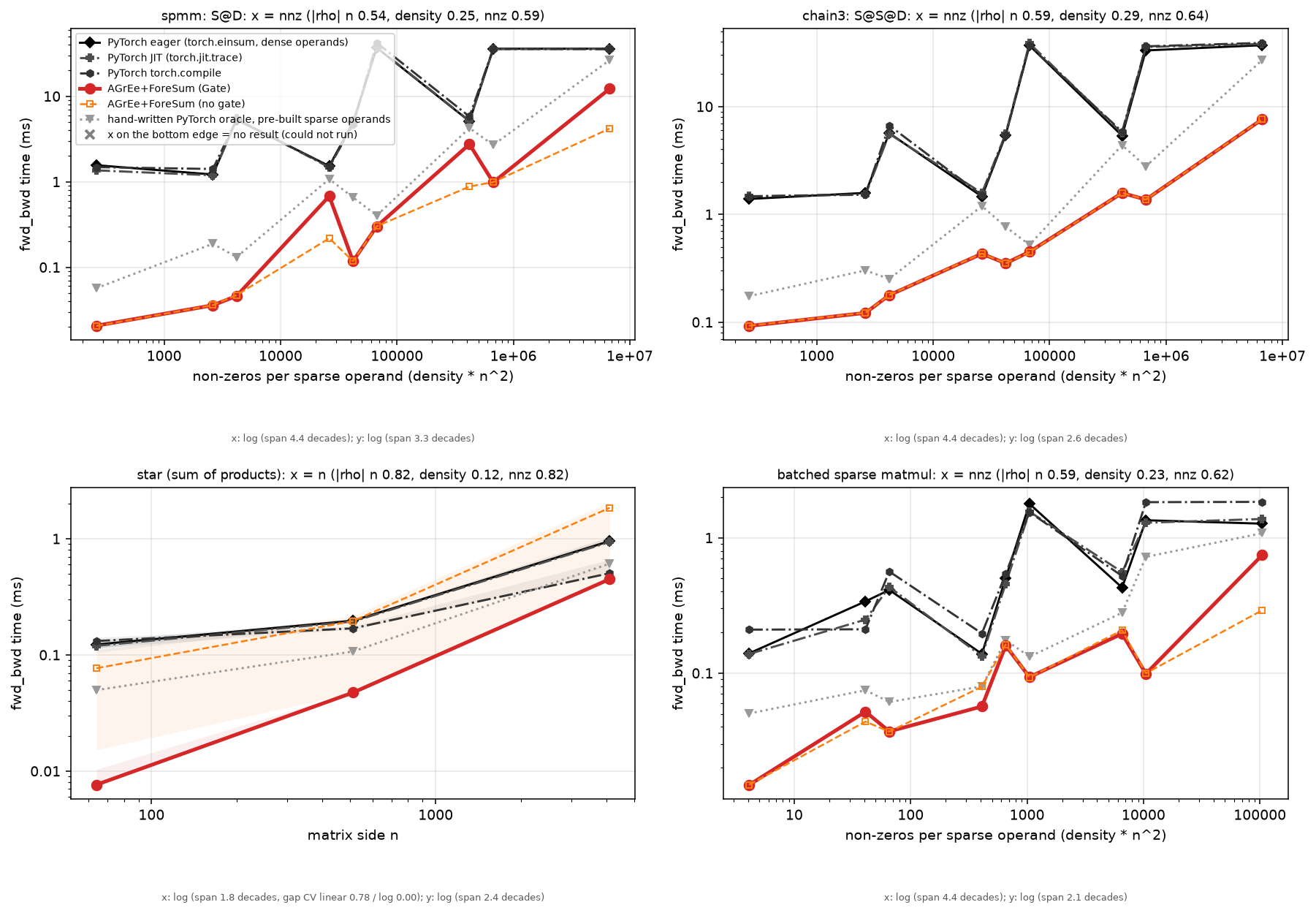}}
				
			}
			
			\subcaption{\label{fig-seconda}Scalability results for both the forward
				and backward step.}
			
		\end{minipage}%
		
		\caption{\label{fig-gruppo}AGrEe +ForESum vs PyTorch, one panel per type
			of query (line = median over the cases sharing x, band = min-max)}
		
	\end{figure}%
	
	In this final set of experiments, we extend the pipeline from the first
	experiment to also introduce an Einstein Sum computation node, while
	keeping all the remaining nodes intact. Even in this case, we consider
	both forward and backwards running times across the four sets of queries.
	We consider the same PyTorch configurations, while pairing those with
	hand-written solutions on PyTorch for directly handling sparse tensors by
	compiling the solution out of our YAML configuration file.
	
	Figure~\ref{fig-prima} clearly shows that, overall, PyTorch run-time
	results clearly do not depend on density factors like our solutions,
	thus only depending on the actual tensor size. This, jointly with the
	requirement to allocate a conspicuous amount of memory for intermediate
	tensors, justifies their overall slowdowns. By looking at the raw data
	from the experiments, it is also clear that, when sparse routes are
	picked, the runtime scales with the number of non-zeros while, in dense
	cases, our solution scales like PyTorch but with a better kernel.
	
	While forward experiments are almost identical with and without the
	pre-gate, their effects are more evident when also considering the
	backward passes. Figure~\ref{fig-seconda} for \(d \geq 0.1\) on star,
	the pre-gate takes the right decision to follow the dense delivery for
	the backward pass, while the forward expert doesn't make that mistake.
	The same pattern appears in the first call, with open 4--26× slower on
	dense-regime cases.
	
	\section{Future work}\label{future-works}
	
	The present work could be further generalised by completely avoiding the
	training cost for estimating the running time of each route while
	exploiting hardware probing for deriving explainable results.
	Preliminary results suggested that doing so requires moving beyond the
	simple theoretical formulation of time complexity by also taking into
	account the actual machine model, which is often undocumented. Probes
	are then required to better understand the cache effects for each kernel
	over different machine architectures for then deriving a general time
	complexity formulation which should be less dependent on parameter
	folding. Future work will also aim to fully test this pipeline for
	performing actual NeSy computations over large and sparse tensors. Last,
	we will parallelise the execution plan by running multiple operators in
	parallel rather than parallelising the execution of each single
	operator, as suggested by our earlier work \autocite{info14030173}.

	\printbibliography

\end{document}